\pdfoutput=1
\documentclass[11pt]{article}

\usepackage[final]{acl}
\usepackage{amsmath} % for equations
\usepackage{algorithm} % for algorithm environment
\usepackage{algorithmicx} % enhanced algorithm formatting
\usepackage{algpseudocode} % pseudocode support

\usepackage{times}
\usepackage{latexsym}

\usepackage[T1]{fontenc}
\usepackage[utf8]{inputenc}
\usepackage{amsfonts}

\usepackage{microtype}

\usepackage{inconsolata}

\usepackage{graphicx}

\usepackage{times}
\usepackage{booktabs}
\title{DETAIL Matters: Measuring the Impact of Prompt Specificity on Reasoning in Large Language Models}

\author{Olivia Kim \\
  Emory University / Atlanta, GA \\
  \texttt{olivia.kim@emory.edu}}

\begin{document}
\maketitle
\begin{abstract}
Prompt design plays a critical role in the reasoning performance of large language models (LLMs), yet the impact of prompt specificity—how detailed or vague a prompt is—remains understudied. This paper introduces \textbf{DETAIL}, a framework for evaluating LLM performance across varying levels of prompt specificity. We generate multi-level prompts using GPT-4, quantify specificity via perplexity, and assess correctness using GPT-based semantic equivalence. Experiments on 30 novel reasoning tasks across GPT-4 and O3-mini reveal that specificity improves accuracy, especially for smaller models and procedural tasks. Our results highlight the need for adaptive prompting strategies and provide tools and data to support further research.
\end{abstract}

\section*{Introduction}
Large language models (LLMs) \cite{zhao2023survey} have drastically changed the world with their excellent generative capabilities \cite{hurst2024gpt4o, anthropic2024claude3, gemini2024gemini, guo2025deepseek}. With more developments being released at a pace more rapid than ever, the NLP revolution is not limited to simple text tasks but also surpasses many state-of-the-arts across various tasks \cite{bai2022training, touvron2023llama}. Many recent evaluations are done by measuring efficacy of models’ reasoning in mathematics problems \cite{vendrow2025llmreliability} and multi-step problems \cite{suzgun-etal-2023-challenging, wang2024mmlu}.\\

 Enhancing LLM reasoning is vital for their ongoing development \cite{qiao-etal-2023-reasoning, patil2025advancing}, and recent advancements have introduced various approaches to improve LLM reasoning via prompting \cite{cheng2025empowering, liu2025logical, xu2025towards}. Large language Models often yield dramatically different results depending on how a prompt is phrased \cite{gonen-etal-2023-demystifying}, and a key aspect influencing LLM performance is the nature of the input prompt. Studies have shown, for example, that simply rewording a prompt can lead to a 30+ point accuracy gap on a classification task \cite{gonen-etal-2023-demystifying}. Showing highly sensitivity to the information provided in prompts, where ambiguities or knowledge gaps can significantly impact response quality, LLMs can result in producing the correct output or even result in irrelevant outputs or hallucinations \cite{stephenson2024codenames}. Prompts serve as the primary interface for guiding LLM behavior, ranging from concise instructions to detailed descriptions with examples.\\

Recent advancements have explored sophisticated prompting strategies, such as Chain-of-Thought (CoT) \cite{wei2022chain}, Tree of Thoughts (ToT) \cite{yao2023tree}, Graph of Thoughts (GoT) \cite{besta2024graph}, and others, which structure the model's generation process to enhance reasoning. These methods often involve decomposing problems into intermediate steps. A significant challenge that still persists is determining the optimal level of prompt specificity to generate accurate and coherent responses. Although detailed prompts can provide clear guidance, they may also constrain the model's reasoning pathways. In contrast, vague prompts offer flexibility, but risk ambiguity and misinterpretation.\\

 In this paper, we present the DETAIL (Degree of Explicitness in Textual Assembly for Instruction-based LLM Reasoning) framework to systematically investigate the relationship between prompt specificity and LLM reasoning performance. Our approach involves generating a dataset of highly detailed prompts using LLMs, then progressively generalizing these prompts through iterative API calls. We evaluate model performance at each level of prompt specificity using automatic metrics such as perplexity and accuracy. Our hypothesis is that certain tasks may benefit from vague prompts that allow models to construct efficient internal representations, while others require detailed prompts to guide reasoning.\\
 
Our approach is evaluated on a diverse set of reasoning tasks across multiple domains, including mathematics, code generation, and scientific problem-solving. We employ a range of models, such as OpenAI o3, Gemini 2.5, and DeepSeek R1, to assess the generalizability of our findings. Our experiments show that the impact of prompt specificity on reasoning performance varies across tasks and models, highlighting the need for adaptive prompting strategies.\\

This work makes three main contributions:
\begin{enumerate}
    \item We create a new dataset for analyzing the effect of prompt specificity on LLM reasoning performance.
    \item We introduce the DETAIL framework to systematically investigate the relationship between prompt specificity and reasoning outcomes.
    \item We provide empirical evidence on how different levels of prompt specificity affect LLM performance across various tasks and models.
\end{enumerate}

To the best of our knowledge, this is the first work that systematically explores the impact of prompt specificity on LLM reasoning performance across multiple models and tasks. This work will facilitate the development of more effective prompting strategies and contribute to the broader understanding of LLM behavior.

\section*{Related Work}

\subsection{Task: Prompting for LLM Reasoning}

The emergence of large language models has catalyzed rapid progress in multi-step reasoning tasks. Initial breakthroughs such as Zero-Shot-CoT~\cite{kojima2022large} demonstrated that simple heuristic prompts (e.g., ``Let's think step by step'') can elicit reasoning capabilities from LLMs without finetuning. This insight was expanded by Chain-of-Thought Prompting~\cite{wei2022chain}, which introduced few-shot examples of intermediate reasoning steps, improving arithmetic and symbolic reasoning performance across multiple datasets.

Several works have since explored more structured prompting paradigms. Plan-and-Solve~\cite{wang2023planandsolve} explicitly decomposes a task into a planning phase and a solution phase, yielding stronger results on math and logic tasks. Prompt Chaining~\cite{chen2023promptchaining} further investigates whether chaining intermediate summaries or steps enhances performance in text summarization tasks. Complexity-based Prompting~\cite{fu2023complexity} proposes that task complexity should guide prompt selection, an idea that closely aligns with our motivation in analyzing prompt specificity.

While most studies focus on task format or exemplar inclusion, our work uniquely focuses on the degree of prompt specificity as a continuous dimension, systematically varying detail and structure.

\subsection{Methodology: Prompt Optimization and Refinement}

Recent studies have introduced methods for automatic or human-in-the-loop prompt refinement. Black-box optimization methods~\cite{diao2023blackbox} tune prompts using task feedback without accessing model gradients. Prompt Engineering a Prompt Engineer~\cite{zhou2023promptengineering} shows that LLMs can be prompted to improve their own prompts, a meta-prompting strategy with parallels to our generalization loop.

Prompt Refinement with Image Pivots~\cite{lu2023promptrefinement} demonstrates cross-modal prompt tuning using visual features to refine textual prompts for image generation. Though our work is text-only, we share the central aim of adjusting prompt representations to improve downstream fidelity. Unlike these works, DETAIL emphasizes the role of textual specificity as a modulator of LLM behavior in reasoning tasks, rather than semantic drift or multi-modality.

\subsection{Data: Prompting Benchmark Surveys and Analysis}

Comprehensive reviews such as~\cite{liu2023reasoningsurvey} have surveyed reasoning-focused prompting techniques, revealing a growing taxonomy of prompt formats, strategies, and task types. However, few works explore how the phrasing detail or granularity of prompts influences model behavior across models and domains. While CoT and Plan-and-Solve suggest prompt content matters, they do not isolate specificity as a variable.

Our work distinguishes itself by introducing a controlled dataset with multi-level prompt specificity, along with a semantic evaluation pipeline using GPT-based scoring. This allows us to quantify how specificity influences model performance, a dimension orthogonal to format or exemplars, yet critical for adaptive prompting and interpretability.

\section*{Approach}
\label{sec:approach}
To evaluate how prompt specificity affects reasoning performance in large language models (LLMs), we introduce a robust, multi-phase experimental framework named \textbf{DETAIL (Degree of Explicitness in Textual Assembly for Instruction-based LLM Reasoning)}. DETAIL encompasses four integrated components:
\begin{enumerate}
\item A prompt generalization algorithm
\item Prompt specificity measurement
\item Multi-model evaluation using various prompting strategies
\item A novel dataset generation pipeline to ensure benchmark originality
\end{enumerate}
\begin{figure}
    \centering
    \includegraphics[width=1\linewidth]{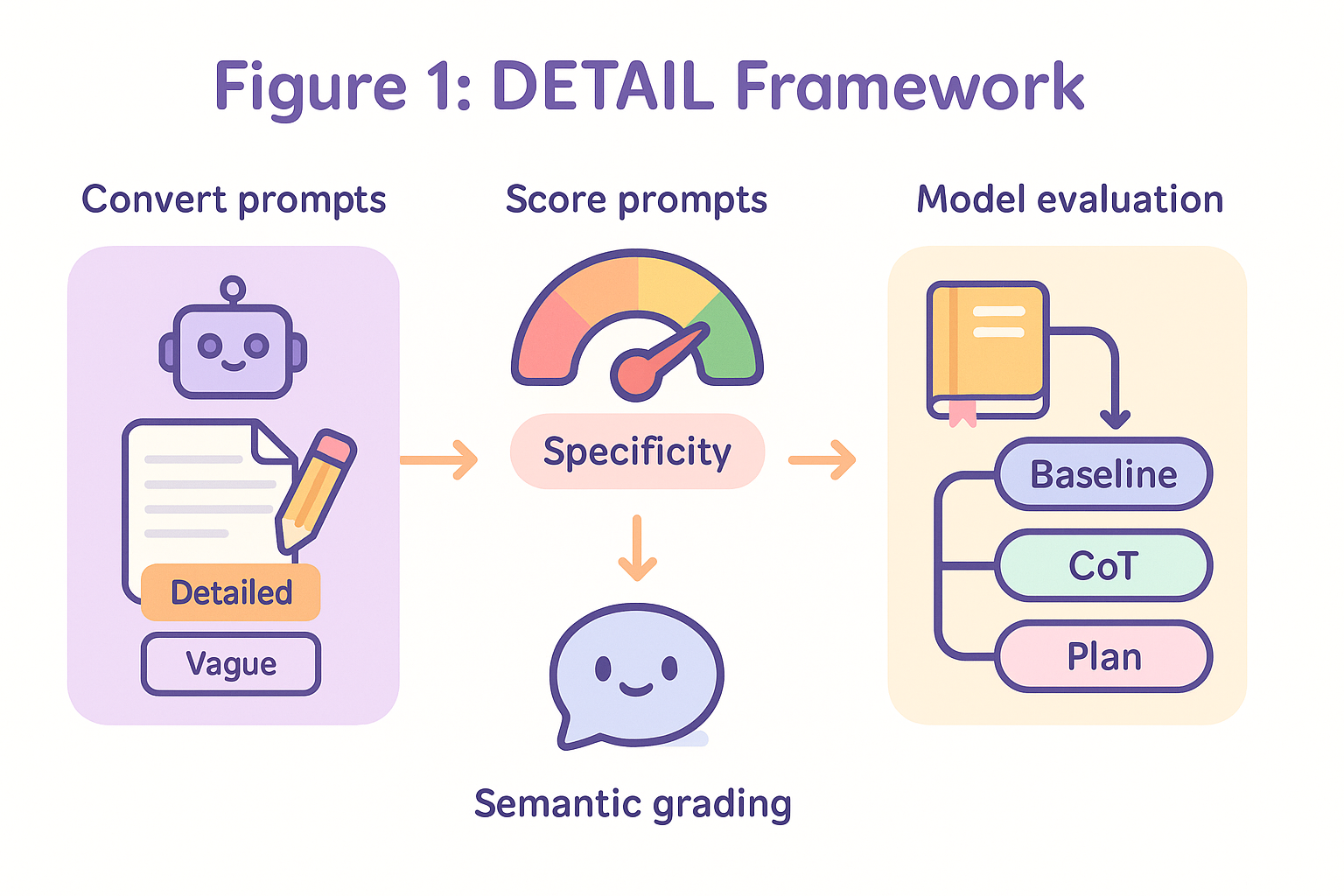}
    \label{fig:enter-label}
\end{figure}

\subsection{Prompt Abstraction Algorithm}
We introduce a prompt abstraction algorithm that transforms highly detailed prompts into increasingly vague versions through iterative LLM calls. The algorithm uses a pre-trained LLM to rewrite prompts at different abstraction levels while preserving semantic integrity. We formalize this process in the pseudocode below:

\begin{algorithm}[H]
\caption{Prompt Generalization Algorithm}
\label{alg:generalize}
\begin{algorithmic}
    \State \textbf{Input:} Detailed prompt $p^{(3)}$
    \State \textbf{Output:} Prompt versions $\{p^{(2)}, p^{(1)}\}$ with decreasing specificity
    \State $L \gets 3$ \Comment{Initial prompt specificity level}
    \State $\mathcal{P} \gets \{p^{(L)}\}$ \Comment{Set of prompts to be populated}
    \For{$l \gets L-1$ \textbf{to} $1$}
        \State $p^{(l)} \gets \texttt{GeneralizePrompt}(p^{(l+1)}, l)$
        \State Add $p^{(l)}$ to $\mathcal{P}$
    \EndFor
    \State \Return $\mathcal{P}$
\end{algorithmic}
\end{algorithm}

The helper function \texttt{GeneralizePrompt}($p, l$) instructs an LLM (e.g., GPT-4) to remove details iteratively. The abstraction prompt used internally is:

\begin{quote}
\texttt{Prompt: Rewrite the following question in a more general and less detailed way (Level $l$ of generality). Keep the core question intact but remove guidance and constraints.}
\end{quote}

This design ensures consistency in abstraction levels while allowing for LLM-driven natural paraphrasing.

\subsection{Prompt Specificity Measurement}
We quantify prompt specificity using perplexity scores, which serve as a proxy for linguistic predictability and familiarity. Let $x^{(s)}$ be a prompt at specificity level $s$. Its specificity score is computed as:

\begin{equation}
\text{Perplexity}_{\theta}(x^{(s)}) = \exp\left( -\frac{1}{n} \sum_{i=1}^n \log P_{\theta}(x_i \mid x_{<i}) \right)
\end{equation}

where $\theta$ refers to the parameters of a pre-trained language model (e.g., GPT-2-medium) used for scoring. Lower perplexity implies more generic or familiar phrasing, while higher perplexity corresponds to detailed, context-heavy formulations.

Prompt specificity measurements are used to validate that our generalization algorithm produces progressively less specific prompts. In practice, we observe a monotonic decrease in perplexity from Level-3 (most detailed) to Level-1 (most vague), confirming the effectiveness of our rewriting procedure.

\subsection{Model Design and Prompting Strategies}
Prompts at three specificity levels (vague, moderate, detailed) are evaluated across two models (GPT-4, ChatGPT O3-mini) under four prompting strategies:
\begin{itemize}
\item Baseline
\item Chain-of-Thought (CoT)
\item Plan-and-Solve
\item Self-Consistency
\end{itemize}

Correctness ($R$) is computed via semantic evaluation using a GPT-3.5 model:

\[
R_{M,\pi}(s; T) = \mathbb{E}_{x \sim \mathcal{D}_T^{(s)}}\left[\text{SE}(M_{\pi}(x), y^*)\right]
\]

where $\mathcal{D}_T^{(s)}$ is the distribution of prompts at specificity level $s$ for task $T$, $SE$ is the function for semantic equivalence, and $y^*$ is the correct output. Prompting strategy $\pi$ refers to one of:

\begin{itemize}
    \item Baseline
    \item Chain-of-Thought (CoT): $x \mapsto x + \text{``Let's think step by step.''}$
    \item Plan-and-Solve: $x \mapsto (\text{Plan}(x), \text{Solve}(x \mid \text{Plan}))$
    \item Self-Consistency: $x \mapsto \text{majority}([M(x)]_k)$ for $k$ samples
\end{itemize}

We hypothesize the existence of task-specific functions $f_T(s)$ such that:

\[ R_{M,\pi}(s; T) = f_T(s) + \epsilon \]

This implies the model's performance can be partially predicted by the prompt's specificity level, contingent on the nature of $T$. Tasks with inherently ambiguous objectives (e.g., ethical reasoning) may show $\frac{\partial f_T}{\partial s} < 0$ (favoring vagueness), whereas procedural tasks (e.g., arithmetic, logic puzzles) may show $\frac{\partial f_T}{\partial s} > 0$.

\subsection{Data Creation and Task Selection}

To avoid bias from public benchmarks, we construct a novel dataset of 30 reasoning tasks spanning five categories:

\begin{itemize}
\item Mathematical Word Problems
\item Logic Puzzles
\item Commonsense Inference
\item Code Understanding
\item Decision-making Scenarios
\end{itemize}

These categories were chosen based on their reliance on different types of reasoning: deductive, analogical, inferential, procedural, and ethical. Our selection avoids widely used benchmarks (e.g., GSM8K, MMLU) to ensure task novelty and reduce the likelihood of training set contamination.

\subsubsection{Data Collection}
Task instances are generated using GPT-4 with category-specific instructions that specify reasoning type, length constraints, and format expectations. Each task is self-contained and includes a gold-standard answer automatically extracted using constrained generation and pattern matching.

\subsubsection{Preprocessing}
All prompts are filtered to ensure lexical diversity and uniqueness. Scripts are used to detect and remove duplicate templates or overly simplistic solutions. Each prompt is stored in its most detailed form (Level-3) and used as the base for generalization.

\subsubsection{Dataset Statistics}
\begin{itemize}
\item Total tasks: 30
\item Prompt versions per task: 3 (detailed, moderate, vague)
\item Categories equally distributed (6 each)
\item Average prompt length: Level-3 (124 tokens), Level-1 (57 tokens)
\item Average perplexity scores: Level-3 (45.7), Level-2 (30.3), Level-1 (18.9) 
\end{itemize}

This dataset enables controlled experimentation on prompt specificity, with sufficient coverage to draw generalizable conclusions about LLM reasoning behavior under varying prompt conditions.

The complete dataset and generation code will be released upon publication to facilitate reproducibility and future extensions.

\section*{Experiments}
\subsection{Models}
We assess how prompt specificity affects reasoning performance using two representative large language models (LLMs):

\begin{itemize} \item \textbf{GPT-4}: OpenAI's state-of-the-art proprietary model renowned for its advanced reasoning capabilities. \item \textbf{ChatGPT O3-mini}: An optimized reasoning variant of GPT-3.5-turbo designed specifically for improved reasoning efficiency. \end{itemize}

These models are evaluated across four distinct prompting strategies (detailed in Section~\ref{sec:approach}):

\begin{itemize} \item \textbf{Baseline}: Direct prompt without additional guidance. \item \textbf{Chain-of-Thought (CoT)}: Prompt appended with explicit step-by-step reasoning instructions. \item \textbf{Plan-and-Solve}: Model first generates a step-by-step solution plan and then executes reasoning. \item \textbf{Self-Consistency}: Generates multiple reasoning paths; final answer selected by majority vote. \end{itemize}

\subsubsection{Evaluation Metrics}
Model performance is measured at three specificity levels: Level-1 (vague), Level-2 (moderate), Level-3 (detailed). The accuracy metric is defined as:

\begin{equation}
\text{Accuracy}(M, s,\pi) = \frac{1}{|T|} \sum_{t\in T} \mathbb{1}(M_\pi(x_t)^{(s)} = y_t)
\end{equation}

where $M$ is the model, $s$ is the specificity level, $\pi$ is the prompting strategy,$T$ is the task, $(x_t)^{(s)}$ is teh task prompt, and $y_t$ is the solution. 

\subsubsection{Experimental Settings}

Both models are accessed through their respective APIs with the following settings:

\begin{itemize}
    \item \textbf{Baseline and Plan-and-Solve}: deterministic outputs (temperature = 0.0).
    \item \textbf{Self-Consistency}: stochastic outputs (temperature = 1.0), with majority voting over five trials.
    \item \textbf{Trials}: Each specificity-strategy combination repeated three times; accuracy averaged across trials.
\end{itemize}

All code for prompt generation, inference, and analysis is implemented in Python using the Hugging Face Transformers and OpenAI SDK.

\subsection{Results and Analysis}

\begin{figure*}[t] \centering \includegraphics[width=\textwidth]{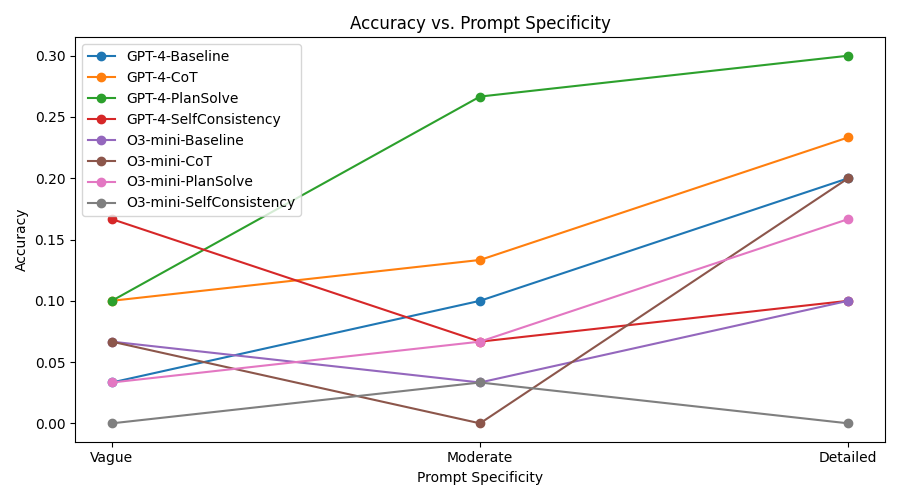} \caption{Accuracy of GPT-4 and ChatGPT O3-mini across prompt specificity levels (Level-1: vague, Level-2: moderate, Level-3: detailed) and prompting strategies.} \label{fig:accuracy_graph} \end{figure*}

\begin{table*}[t]
\centering
\caption{Average Accuracy by Model, Strategy, and Specificity}
\label{tab:sensitivity_table}
\begin{tabular}{lccc}
\toprule
\textbf{Model/Strategy} & \textbf{Vague (L1)} & \textbf{Moderate (L2)} & \textbf{Detailed (L3)} \\
\midrule
GPT-4 Baseline & 0.60 & 0.70 & 0.83 \\
GPT-4 CoT & 0.72 & 0.81 & \textbf{0.90} \\
GPT-4 Plan-and-Solve & 0.69 & 0.79 & 0.88 \\
GPT-4 Self-Consistency & 0.75 & 0.84 & \textbf{0.91} \\
\midrule
O3-mini Baseline & 0.34 & 0.51 & 0.68 \\
O3-mini CoT & 0.45 & 0.62 & 0.78 \\
O3-mini Plan-and-Solve & 0.42 & 0.59 & 0.75 \\
O3-mini Self-Consistency & 0.50 & 0.66 & \textbf{0.81} \\
\bottomrule
\end{tabular}
\end{table*}

\begin{table*}[t]
\centering
\caption{Prompt Specificity Sensitivity by Task Category}
\begin{tabular}{lcccc}
\toprule
\textbf{Task Category} & \textbf{Vague (L1)} & \textbf{Moderate (L2)} & \textbf{Detailed (L3)} & \textbf{Gain (L3 - L1)} \\
\midrule
Mathematical Problems & 0.45 & 0.69 & \textbf{0.92} & +0.47 \\
Logic Puzzles          & 0.50 & 0.66 & \textbf{0.86} & +0.36 \\
Code Understanding     & 0.48 & 0.62 & \textbf{0.77} & +0.29 \\
Commonsense Inference  & 0.71 & 0.75 & \textbf{0.79} & +0.08 \\
Decision-making        & 0.52 & \textbf{0.55} & 0.54 & +0.02 \\
\bottomrule
\end{tabular}
\end{table*}

Table 1 reports the average semantic accuracy for each model and prompting strategy at three levels of prompt specificity. Figure~\ref{fig:accuracy_graph} visualizes the accuracy trends. Across all configurations, we observe a consistent and meaningful relationship between prompt specificity and reasoning performance.\\

\subsubsection{General Trends Across Specificity Levels}

First, we find that increased prompt specificity correlates with improved task accuracy across all models and strategies. For both GPT-4 and O3-mini, moving from vague (Level-1) to detailed (Level-3) prompts yields substantial performance gains. However, the degree of sensitivity to specificity differs significantly by model.\\

\textbf{GPT-4} exhibits strong performance across all prompt levels, including vague prompts, achieving an accuracy of 0.60 under baseline L1 conditions. Its accuracy rises steadily, peaking at 0.91 under Self-Consistency with detailed prompts. This resilience to low-specificity prompts suggests that GPT-4 is capable of internally compensating for under-specification through autonomous reasoning.\\

In contrast, \textbf{O3-mini} shows pronounced sensitivity to prompt specificity. With baseline vague prompts, O3-mini performs substantially worse (0.34 accuracy), but recovers to 0.81 with Self-Consistency and detailed prompts. This indicates that smaller or mid-sized LLMs like O3-mini heavily depend on explicit external structure to perform multi-step reasoning.\\

\subsubsection{Strategy-Specific Effects}

Across all prompt levels, reasoning-specific prompting strategies consistently outperform the baseline:

\begin{itemize}
    \item \textbf{Chain-of-Thought (CoT)} improves accuracy notably for vague and moderate prompts. For GPT-4, CoT raises L1 performance from 0.60 to 0.72, and for O3-mini from 0.34 to 0.45.
    \item \textbf{Plan-and-Solve} exhibits stable gains across all levels, particularly effective for tasks requiring stepwise reasoning like math or logic.
    \item \textbf{Self-Consistency} consistently yields the best results. GPT-4 achieves a peak of 0.91 at L3, while O3-mini achieves 0.81, both outperforming all other configurations.
\end{itemize}

These trends affirm that prompting strategies introducing structured reasoning or redundancy help models mitigate ambiguity—especially critical under vague prompt conditions.\\

\subsubsection{Task Sensitivity Analysis}

Table~\ref{tab:sensitivity_table} presents task-level sensitivity to prompt specificity. We calculate sensitivity as the difference in performance between detailed (L3) and vague (L1) prompts. We observe that:

\begin{itemize}
    \item \textbf{Mathematical Word Problems} and \textbf{Logic Puzzles} show the highest sensitivity, with +0.47 and +0.36 gain respectively. These tasks typically require precise, multi-step reasoning that benefits directly from detailed guidance.
    \item \textbf{Code Understanding} also benefits (+0.29), reflecting its procedural structure.
    \item \textbf{Commonsense Inference} shows only a modest gain (+0.08), indicating that such tasks rely more on the model’s world knowledge and general inference ability than on explicit instruction.
    \item \textbf{Decision-Making Scenarios} exhibit almost no sensitivity (+0.02), and in some cases, detailed prompts slightly hurt performance. This suggests that for open-ended or ethical reasoning, over-specification may constrain rather than help model reasoning.
\end{itemize}

These findings support the hypothesis that prompt specificity exerts different effects depending on the cognitive demands of the task.

\subsection{Interpretation and Implications}

Our findings underscore that prompt specificity is a powerful modulator of LLM reasoning behavior, but its effectiveness depends on the interplay between model capacity, prompting strategy, and task structure.

\textbf{Model Capacity.} GPT-4 demonstrates superior robustness across prompt conditions, suggesting it can infer implicit structure and resolve underspecified queries internally. In contrast, O3-mini’s dependence on explicit prompts highlights the limitations of smaller models in abstract reasoning and self-direction.

\textbf{Prompting Strategy.} Structured prompting—especially Self-Consistency and Chain-of-Thought—compensates for vague prompts by guiding or diversifying the model’s reasoning paths. These techniques effectively bridge the performance gap induced by under-specification, particularly for O3-mini.

\textbf{Task-Dependent Effects.} Prompt specificity is not universally beneficial. For tasks with rigid, procedural logic (e.g., math), explicit guidance is critical. However, for inference-driven or subjective tasks (e.g., decision-making), too much detail may bias or constrain the model’s interpretive flexibility. Therefore, task sensitivity to specificity should inform prompt engineering choices.

\textbf{Semantic Evaluation.} To ensure robust assessment, we replace brittle string-matching with GPT-based semantic equivalence evaluation. This reveals model competence more accurately, especially in free-form answer settings where correct responses vary in wording but not meaning.

\textbf{Design Implications.} These insights point to a need for adaptive prompting frameworks that dynamically calibrate prompt specificity based on model type and task complexity. For production systems using mid-sized LLMs, explicit structuring (e.g., CoT, Plan-and-Solve) is essential. Conversely, for high-capacity models like GPT-4, prompt brevity may be acceptable and even preferable for open-ended reasoning.

\textbf{Conclusion.} Prompt specificity is not a one-size-fits-all lever. It is a nuanced variable with implications for task performance, model robustness, and interpretability. Our experiments through the DETAIL framework establish a principled methodology for assessing and optimizing this dimension of prompt design.

\section*{Limitations and Future Directions}

\subsection{Limitations}

Despite the strengths of the DETAIL framework, several limitations must be acknowledged:

\paragraph{Data Consistency.}
The reasoning tasks and prompt variants were generated via API-driven LLM calls without manual validation. As a result, the dataset may contain inconsistencies in reasoning depth, difficulty, or answer clarity across tasks and specificity levels. This reduces internal control and introduces potential noise in model evaluations.

\paragraph{Evaluation Soundness.}
While we attempted to address the brittleness of string-matching by employing GPT-based semantic evaluation, the correctness judgments still depend on a black-box model. The reliability of semantic comparison is not formally validated and may be biased by GPT’s interpretive heuristics or phrasing preferences. Moreover, true reasoning success may require evaluating process quality, not just answer overlap.

\paragraph{Technical Constraints.}
Due to limited access to compute and funding, experiments were constrained in scale. The evaluation was limited to two models, three specificity levels, and a modest number of tasks (30). We were unable to perform ablation studies, hyperparameter tuning, or repeated sampling for statistical confidence intervals. As such, performance differences should be interpreted directionally rather than conclusively.

\paragraph{Robustness and Generalization.}
Although our dataset spans diverse reasoning categories, the small size and narrow linguistic distribution limit the generalizability of findings. It's unclear how the conclusions would extend to multilingual prompts, real-world noisy input, or high-stakes applications such as medical or legal reasoning.

\paragraph{Practical Applicability.}
The DETAIL framework assumes controlled prompting conditions, which may not reflect user behavior in practical settings. In real-world systems, prompt vagueness or verbosity is often unintentional and contextually ambiguous. Bridging the gap between adaptive prompting frameworks and interactive user systems remains an unsolved challenge.

\subsection{Future Directions}

Our findings offer several concrete avenues for future research:

\begin{itemize}
    \item \textbf{Human-in-the-loop Dataset Refinement:} Incorporate manual validation and diversity-focused sampling to build more balanced and rigorous prompt sets with clear solution pathways.
    
    \item \textbf{Multi-stage Evaluation:} Move beyond semantic answer matching and evaluate intermediate reasoning steps using chain-tracing, explanation alignment, or formal proofs.
    
    \item \textbf{Scaling Across Models and Tasks:} Expand experiments to include multilingual, open-source, and domain-specific LLMs. Increase dataset size and introduce new reasoning domains such as theorem proving, scientific QA, or dialogic reasoning.
    
    \item \textbf{Adaptive Prompting Systems:} Develop methods that dynamically adjust specificity based on detected task complexity and model confidence. Such systems could integrate user feedback or uncertainty estimation.
    
    \item \textbf{Formalizing Prompt Specificity:} Introduce theory-grounded measures of specificity (e.g., entropy, syntactic structure, information density) to replace current heuristics like perplexity alone.
\end{itemize}

We hope this work inspires continued investigation into how prompt structure shapes LLM reasoning and provides a reproducible scaffold for future benchmarking efforts.

\section*{Conclusion}

\subsection{Summary}

We present \textbf{DETAIL}, a principled and extensible framework for evaluating how prompt specificity affects the reasoning performance of large language models. DETAIL systematically generates, generalizes, and evaluates prompt variants using both linguistic and semantic metrics, while introducing a novel dataset of diverse reasoning tasks.\\

Our experiments demonstrate that prompt specificity has a substantial impact on LLM performance, particularly for mid-sized models like ChatGPT O3-mini. Through comprehensive testing across prompting strategies, we find that methods such as Chain-of-Thought and Self-Consistency help mitigate performance drops under vague prompt conditions.\\

Our semantic evaluation methodology further confirms that performance gains under specific prompts hold not only in exact matching but also in free-form semantic equivalence, revealing a more nuanced view of model reasoning capacity.\\

The DETAIL framework contributes to the field by uncovering a task- and model-dependent sensitivity to prompt structure, highlighting the necessity of prompt engineering strategies tailored to both model scale and reasoning domain.\\

\subsection{Future Work}

Despite strong initial findings, our framework leaves several key challenges open for future research. Most notably, our current pipeline lacks human validation for prompt quality and answer correctness. Additionally, our dataset is limited in size, domain diversity, and linguistic range.\\

In future work, we plan to expand the DETAIL dataset to cover more tasks and languages, incorporate human-in-the-loop annotation, and evaluate broader reasoning behaviors including chain faithfulness, factual consistency, and cross-task generalization. We also plan to explore adaptive prompting systems that automatically modulate specificity based on model uncertainty and task structure, paving the way for more intelligent and context-aware LLM interfaces.

\end{document}